\def\year{2020}\relax
\documentclass[letterpaper]{article} 
\usepackage{aaai20}  
\usepackage{times}  
\usepackage{helvet} 
\usepackage{courier}  
\usepackage[hyphens]{url}  
\usepackage{graphicx} 
\usepackage{graphicx}  
\usepackage{amsmath}
\usepackage{amsfonts}
\usepackage{ wasysym }
\usepackage{tabularx}
\usepackage{booktabs}
\usepackage[ruled ]{algorithm2e}

\newcommand{\citet}[1]{\citeauthor{#1}~\shortcite{#1}}
\newcommand{\citep}{\cite}
\newcommand{\citealp}[1]{\citeauthor{#1}~\citeyear{#1}}

\title{The Efficiency of Human Cognition Reflects \\ Planned Information Processing}
\author{
Mark K. Ho,\textsuperscript{\rm 1}
David Abel,\textsuperscript{\rm 2}
Jonathan D. Cohen,\textsuperscript{\rm 1}  \\
\Large
\textbf{
Michael L. Littman,\textsuperscript{\rm 2} 
Thomas L. Griffiths\textsuperscript{\rm 1} }\\
\textsuperscript{\rm 1} Department of Psychology, Princeton University, Princeton, NJ \\
\textsuperscript{\rm 2} Department of Computer Science, Brown University, Providence, RI \\
\texttt{\{mho, jdc, tomg\}@princeton.edu} \\ \texttt{\{dmabel, mlittman\}@cs.brown.edu}
}

\begin{document}

\maketitle

\begin{abstract}
Planning is useful. It lets people take actions that have desirable long-term consequences. But, planning is hard. It requires thinking about consequences, which consumes limited computational and cognitive resources. Thus, people should plan their actions, but they should also be smart about how they deploy resources used for planning their actions. Put another way, people should also ``plan their plans''. Here, we formulate this aspect of planning as a meta-reasoning problem and formalize it in terms of a recursive Bellman objective that incorporates both task rewards and information-theoretic planning costs. Our account makes quantitative predictions about how people should plan and meta-plan as a function of the overall structure of a task, which we test in two experiments with human participants. We find that people's reaction times reflect a planned use of information processing, consistent with our account. This formulation of planning to plan provides new insight into the function of hierarchical planning, state abstraction, and cognitive control in both humans and machines.
\end{abstract}

\section{Introduction}
Suppose you have to catch a flight in three hours, and you are in your bedroom packing. How would you plan your next move? You might reason, ``I will finish packing my bag. Then, I will get a car to take me to the airport.'' This plan seems intuitive and straightforward. However, many details have been left out: Which taxi or ride-sharing company will you use? How are you going to get from security to your gate? If you reach the gate and get hungry, can you pick up a snack in time to still make your flight? Clearly, you did not imagine every possible contingency from now until your flight departs. Instead, you sketched out a partial plan that considers the information most relevant to your current circumstance, and delayed thinking about other details to when they become more relevant. For instance, rather than thinking of a detailed path from your bedroom to the flight gate right now, you might \textit{plan to later think about} the route to the gate \textit{once you get through airport security}.

As this example suggests, human decision making not only involves planning one's actions, but also planning one's plans. But why plan one's plans? Why not just plan one's actions? We propose that planning one's plans emerges from two aspects of human decision making. First, plans can include different details at different times and change as time moves on. Second, representing detailed plans is itself costly (in terms of time and memory). Thus, people should consider what details they include in a plan and when to include those details. That is, they should plan their plans.

Here, we develop and empirically evaluate this idea in humans. We first discuss how planning one's plans relates to existing ideas in psychology and machine learning. Then, we formalize the relation between planning one's plans and the cost of planning by defining a general Bellman objective that includes both task rewards and information-theoretic planning costs. We then discuss several qualitative features of solutions to this objective. Finally, we report the results of two new human experiments that compare participant reaction times during problem solving to the predictions of our model and alternatives. These empirical findings support our normative account of planning one's plans and raise new questions about the nature of boundedly rational decision making in both people and machines.

\section{Background}
Put simply, planning involves finding a good sequence of actions given a particular problem representation~\cite{Newell1958}. In computer science, many approaches have been developed to facilitate planning. These include classical planning methods such as depth-limited search~\cite{korf1990}, heuristic search~\cite{pearl1984}, and Monte-Carlo tree search~\cite{Abramson1987} as well as the use of data structures like hierarchies~\cite{sacerdoti1974planning,Kaelbling2010}, temporal abstractions~\cite{mcdermott1998pddl}, and state abstractions~\cite{givan2003equivalence}. At the same time, psychologists have long recognized that people also rely on heuristics~\cite{tversky1974judgment,gigerenzer1996reasoning} and use abstractions to organize their thoughts and behaviors~\cite{Lashley1951,Rosenbaum1984,Solway2014}.

But why do people use heuristics and abstractions, and why do we build these structures into our algorithms to use? A simple reason is that na\"{i}ve planning is prohibitively costly~\cite{bellman1957dynamic,littman1995complexity}, so such aids focus limited computational resources on the most important, urgent, or relevant parts of a problem. This observation motivates the following question: What if usefully and efficiently representing plans were an explicit goal of making decisions over time?

To develop a model of adaptively planning with the costs of planning in mind (i.e., planning one's plans), we draw on ideas from several lines of research. One is work on boundedly optimal intelligence~\cite{Russell1995}, anytime algorithms~\cite{horvitz1990computation,dean1988analysis}, and human rational meta-reasoning~\cite{Griffiths2015}, which articulates the value of managing computational costs during decision making. Another is work on the psychology of intertemporal choice~\cite{berns2007} and representation~\cite{trope2003temporal}, which studies how people's construal of different aspects of the world changes as a function of distance, time, and context. Finally, we draw on formal tools from information-theoretic approaches to bounded rationality~\cite{Tishby2011,Rubin2012,Ortega2013}, which provides a general framework for characterizing the relationship between environmental rewards and decision-making costs.

\section{Planned Information Processing}
We begin by formalizing the objective of planning one's plans in terms of planned information processing. To model how an agent should plan at different points in time given a task and a cost of planning, we treat the decisions of what to plan and when to plan as a sequential decision-making problem~\cite{puterman1994markov}. To capture a general, architecture-agnostic notion of planning costs, we use a formulation of partial planning and information-theoretic action costs~\cite{Tishby2011,Rubin2012,Ortega2013}. Finally, at the end of this section and in the supplementary materials, we present a gradient-based algorithm for solving this objective.

\subsection{Markov Decision Processes}
Sequential decision-making can be described in terms of a Markov Decision Process (MDP)~\cite{puterman1994markov}. A discrete MDP~$M$ is a tuple $\langle S, A, T, R, \gamma \rangle$, where $S$ is the state space; $A$ is the action space; $T: S \times A \rightarrow \Delta(S)$ is a transition function that defines a probability distribution over next states $s' \in S$ having taken action $a \in A$ in state $s \in S$; $R: S \times A \times S \rightarrow \mathbb{R}$ is a reward function that defines agent payoffs; and $\gamma \in [0, 1)$ is a discount rate.\footnotemark
\footnotetext{$\Delta(X)$ denotes the simplex over discrete elements $x \in X$.} 

An agent's \textit{policy} describes its behavior in an MDP. Formally, a policy $\pi: S \rightarrow \Delta(A)$ is a mapping from states $s \in S$ to probability distributions over actions $a \in A$. A policy combined with a transition function defines how an agent will move within the state space over time. Additionally, the \textit{normalized discounted occupancy} of a policy $\pi$ in MDP $M$ is $\rho^{\pi}(s) \propto \sum_{t=0}^{\infty} \gamma \text{Pr} \{s = s_t \}$.

We are interested in the \textit{value function} for a particular policy, $V^\pi:~S~\rightarrow~\mathbb{R}$, which is the expected discounted cumulative reward that an agent receives by following $\pi$ from state $s$ onward. We are especially interested in the \textit{optimal value function} defined by the unique fixed point of the Bellman equation~\cite{bellman1957dynamic}, where for all $s \in S$:
\begin{equation}
V^*(s) = \max_{a \in A} \sum_{s' \in S} T_{s, s'}^a [ R_{s, s'}^a + \gamma V^*(s') ].
\label{eq:bellman}
\end{equation}

The optimal value function describes the best that an agent can expect to do in terms of maximizing future discounted rewards from each state. It is also useful to define the optimal state-action value function $Q^*(s, a) = \sum_{s' \in S}T_{s, s'}^a [ R_{s, s'}^a + \gamma V^*(s') ]$, for all $s \in S, a \in A$. Finally, an optimal policy $\pi^*$ is any policy $\pi$ such that $V^{\pi}(s) = V^*(s)$ for all $s \in S$.

\begin{figure*}[t]
    \centering
    \includegraphics[width=.85\textwidth]{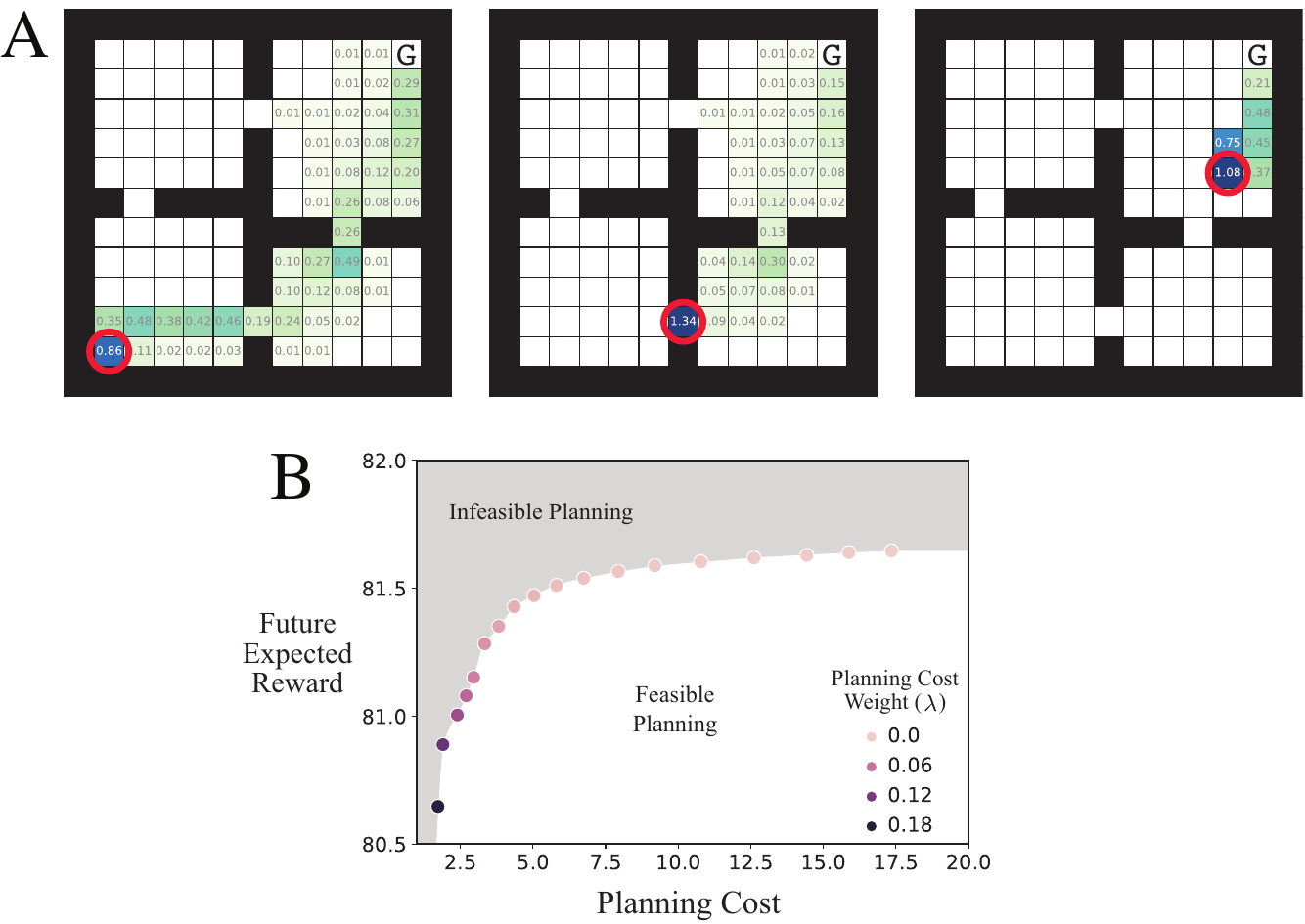}
    \caption{Results on Four Rooms Domain. (A) \textit{Partial Planning Costs}. Shown are the planning costs associated with each simulated state $\tilde{s}$ from three ground states $s$ ($\tilde{s} = s$ are circled in red). Numbers correspond to the KL-divergence of the partial plan from the default plan at each simulated state. The total computational cost for a state is the sum of divergences over the simulated state space (see Equation~\protect\ref{eq:infocost}). Higher values occur at states where the decision is relevant to the current decision, such as the entrances to doors. Values less than $0.005$ are not shown. \texttt{G} denotes the goal. (B)~\textit{Reward-Planning Pareto frontier}. Planning to plan involves both task rewards and planning costs. Depending on the particular task, certain combinations of task rewards and planning costs may be infeasible. Optimizing Equation~\ref{eq:plantoplan} for any particular $\lambda$ finds a point where expected rewards cannot be increased without increasing planning costs and planning costs cannot be decreased without decreasing rewards. Plotted are the planning costs and future value (without planning costs) from the lower-left state in Four Rooms after running our algorithm for a range of $\lambda$ values. The induced Pareto frontier is analogous to an information-theoretic rate-distortion tradeoff~\protect\cite{shannon1948mathematical,Tishby2011}. 
    }
    \label{fig:fourrooms}
\end{figure*}

\subsection{Partial Planning via Soft Planning}
In an MDP, \textit{planning} corresponds to finding a policy that yields high value by doing computations over a \textit{model} of the task. For instance, an optimal policy $\pi^*$ generates plans that perfectly maximize value. But, we may not always want a perfect plan. Rather, since perfection is costly, it is often useful to express imperfect \textit{partial plans}.

We introduce two ideas to formalize partial plans. First, we distinguish between the ground MDP, $M$, and a ``simulated'' MDP, $\tilde{M}$. Here, we focus on the relationship between ground states $S$ and simulated states $\tilde{S}$, while also assuming that $\tilde{M} = M$. This distinction allows us to express different quantities (e.g., action probabilities, expected values) over the same simulated state space but from the perspective of different ground states. For example, we denote the planned probability of taking action $a$ at simulated state $\tilde{s}$ from ground state $s$ as $\tilde\pi(a \mid \tilde{s} ; s)$. In the airport example, you are simulating what you would do at the airport, $\tilde{s}$, from your bedroom, $s$. Note the special case of $\tilde\pi(a \mid s ; s)$ (i.e., $\tilde{s} = s$), which defines the actions an agent plans on taking at their current state.

Second, we introduce a \textit{soft planning parameterization} of partial policies that controls the allocation of planning to different simulated states. Formally, an \textit{inverse temperature assignment} from state~$s$, $\beta(\cdot ; s):~\tilde{S}~\rightarrow~\mathbb{R}_{\geq 0}$, assigns a positive real inverse temperature to each simulated state $\tilde{s} \in \tilde{S}$. Given this assignment, we define soft-Bellman equations over simulated states $\tilde{s} \in \tilde{S}$ from a ground state $s$:
\begin{equation}
\tilde{\pi}^{\beta} (a \mid \tilde{s} ; s) \propto \exp \bigg\{ \tilde{Q}^{\beta}(\tilde{s}, a; s) \beta(\tilde{s} ; s ) \bigg\},
\label{eq:pp0}
\end{equation}
\begin{equation}
\tilde{V}^\beta (\tilde{s}; s) = 
\sum_{a} \tilde{\pi}^{\beta}(a \mid \tilde{s}; s) \tilde{Q}^{\beta}(\tilde{s}, a; s),
\label{eq:pp1}
\end{equation}
\begin{equation}
\tilde{Q}^{\beta}(\tilde{s}, a; s) = \sum_{\tilde{s}'} T_{s, s'}^a
\bigg[ 
R_{s, s'}^a + \gamma \tilde{V}^\beta(\tilde{s}'; s)
\bigg].
\label{eq:pp2}
\end{equation}
Intuitively, the inverse temperature assignment captures how much attention is paid at each simulated state when constructing a partial plan. Larger inverse temperatures entail more attention at a particular state, and the interaction of temperatures induces a partial plan $\tilde{\pi}^\beta(\cdot \mid \cdot; s)$.\footnotemark

\subsection{Information-Theoretic Planning Costs}
How can we quantify the cost of a partial plan? Our proposal borrows the idea of information theoretic costs for actions~\cite{Tishby2011,Rubin2012,Ortega2013} and applies it to simulated (planned) actions. For instance, we can define the cost of a plan $\tilde{\pi}$ based on sum of Kullback-Leibler (KL) divergences, denoted $D_{\text{KL}}$, from a default policy $\bar{\pi}$ at each state:
\begin{equation}
C(\tilde{\pi}, \bar{\pi}) 
= \sum_{\tilde{s} \in \tilde{S}}
D_{\text{KL}}\big[\tilde{\pi}(\cdot \mid \tilde{s}) || \bar{\pi}(\cdot \mid \tilde{s}) \big],
\label{eq:infocost}
\end{equation}
where $D_{\text{KL}}[p || q] = \sum_{x} p(x) \log(\frac{p(x)}{q(x)})$ for distributions $p$ and $q$ with the same support~\cite{Cover1991}. Here, we set $\bar{\pi}$ to be the uniform distribution at all states as it makes few assumptions and is justified by previous work~\cite{Ortega2015}. However, our formulation straightforwardly accommodates task-specific default policies.

Specifying planning costs as an information theoretic quantity has conceptual and practical advantages. First, planning costs can be characterized independently of an agent's specific representation of a task. Second, it can be interpreted as the minimal cost in bits of using the old plan $\bar{\pi}$ to represent the action distributions in the new plan $\tilde{\pi}$~\cite{Cover1991}, which captures intuitions about planning as a costly process of information transformation and control. Finally, the cost term is differentiable with respect to the policy probabilities.

\footnotetext{Previous work has interpreted the soft-maximization in Eq.~\protect\ref{eq:pp0} in terms of the sampling process of a bounded rational decision-maker, where inverse temperature reflects noise~(\protect{\citeauthor{Train2003}, \citeyear{Train2003}, p48; \citeauthor{Ortega2015}, \citeyear{Ortega2015}}). Here, we mainly treat soft-planning as a way to parameterize partial policies and leave a more in-depth analysis of the sampling interpretation of this model for future work.}

\subsection{Planned Information Processing Bellman Objective}

Given our formalization of partial planning and information theoretic planning costs, we can now define the objective of planning one's plans. Put simply, we nest partial planning (Equations~\ref{eq:pp0}--\ref{eq:pp2}) inside a meta-planning problem that includes planning costs (Equation~\ref{eq:infocost}):

\begin{align}
\begin{split}
{V}^*_\lambda(s) = 
\max_{\beta(\cdot ; s)}
\bigg\{
\sum_{a \in A}
\Big[
\tilde{\pi}^{\beta}(a \! \mid \! s; \! s)
Q^*_\lambda(s, a) 
\Big]
\!-\! \lambda C(\tilde{\pi}^{\beta}, \bar{\pi})
\bigg\},
\label{eq:plantoplan}
\end{split}
\end{align}
where $Q^*_\lambda(s, a) = \sum_{s' \in S}T_{s, s'}^a \Big[R_{s, s'}^a + \gamma {V}^*_\lambda(s')\Big]$.



This equation extends the original Bellman equation (Equation~\ref{eq:bellman}) in two ways. First, ground actions are no longer directly chosen. Instead, a distribution over actions is induced at the current state $s$ by influencing the computations of the planner with $\beta(\cdot; s)$. 

Second, Equation~\ref{eq:plantoplan} takes into account the immediate cost of planning via the term $\lambda C(\tilde{\pi}^\beta, \bar{\pi})$, where $\lambda \in \mathbb{R}$ is a planning cost weight. Note that this formulation also includes planning opportunities and costs \textit{in the future} via the recursive nature of the Bellman equation. To understand the significance of this, consider again the airport example: While at home packing, you only vaguely consider how you will get from security to your gate. This is not because it does not matter; if you never make it to your gate and miss your flight, there is no point in packing. Rather, packing does not require you to think about the details of navigating the airport until later. In short, Equation~\ref{eq:plantoplan} expresses how an agent should partially plan to act in the moment, taking into account that they will engage in planning in the future.

\subsection{Optimal Planned Information Processing}
To solve for the meta-planning objective in Equation~\ref{eq:plantoplan}, we implemented a gradient-based algorithm that solves for the optimal inverse temperature assignments (i.e., $\beta^*$) given an MDP $M$, planning cost weight $\lambda$, and default policy $\bar{\pi}$.

To understand optimal meta-planning and how it relates to previous work on hierarchical planning and learning, we ran our algorithm on the Four Rooms domain with deterministic transitions~\cite{sutton1999between}. An absorbing goal state worth $+100$ points was placed in the upper-right corner. Step costs were included ($-.1$ points) and the discount rate was set to $.99$, with $\lambda = 0.01$. Planning iterations were chosen such that value iteration would converge ($H = 100$), while meta-planning parameters were solved using the Adam optimizer~\cite{kingma2014adam} for $N = 200$ iterations (see supplementary materials for details on the algorithm).

The optimization finds an inverse temperature assignment $\beta^*$ that yields a partial plan from each ground state $s$---i.e., $\tilde{\pi}^{\beta^*}(\cdot \mid \cdot ; s)$. The cost of each partial plan is determined by the entire simulated state space (Equation~\ref{eq:infocost}), but we can examine the contribution of the planned actions at each simulated state $\tilde{s}$ to identify which ones are prioritized, shown as colors in Figure~\ref{fig:fourrooms}a. Two salient features emerge in the Four Rooms Domain. First, at any point in time, ``doorways'' leading to the goal are prioritized because the quality of the current decision depends on planning to go through the door and proceeding towards the goal. Second, actions at states that are closer along the path are represented in greater detail since they are more relevant to simulating the value of the current decision. 

Additionally, Equation~\ref{eq:plantoplan} implicitly defines a task-specific \textit{Pareto frontier} where task rewards cannot be improved without worsening planning costs and vice versa. Since our algorithm seeks to find points that jointly maximize rewards and minimize planning costs, we can identify this curve for the Four Rooms task by running our algorithm with a range of $\lambda$ values. Figure~\ref{fig:fourrooms}b shows the plot of this curve at the lower-left state, which divides the space of possible task rewards and planning costs into feasible and infeasible combinations of rewards and planning costs.

\begin{figure*}[!ht]
    \centering
    \includegraphics[width=.9\textwidth]{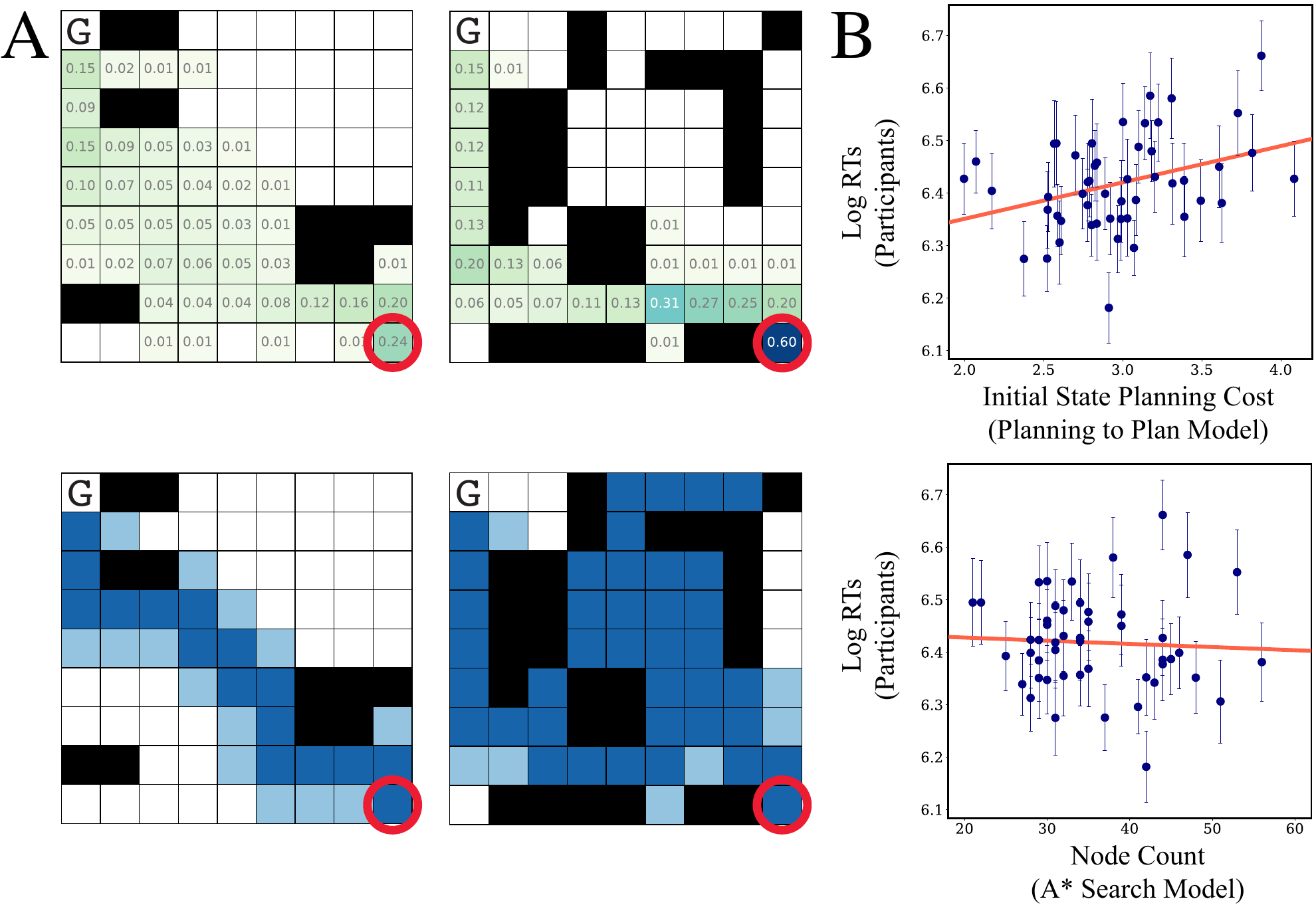}
    \caption{
    Experiment 1 examples and results. (A) Planned information processing (upper pair of examples) versus A$^*$ search (lower pair) on two of 50 mazes from Experiment~1. Partial plans at the initial lower-right state (red circles) are more specific (and costly) at future states relevant to the current decision, represented by darker green (top row). A$^*$ can plan with heuristics (e.g., Manhattan distance to goal) but does not adaptively allocate planning based on task structure (bottom row; dark and light blue are explored states and candidate states, respectively). If people are only planning, and not planning to plan, initial-state response times will reflect processes like A$^*$ and not partial planning. (B) Mean log-normalized RTs for the 50 grids used in the task as a function of partial planning costs (top) and A$^*$ Node Count. Red lines are regression lines for mean log RTs by item and model predictions (Top: $R^2 = 0.12$, $p < 0.05$; Bottom: $R^2 = 0.003; p = 0.71$).
    Error bars represent standard errors.
}
    \label{fig:exp1}
\end{figure*}

\section{Experiment 1: Parametric Mazes}
Our account makes quantitative predictions about how people should flexibly plan information processing on a task. To test these predictions, we examined how long it took people to make decisions when navigating through a set of parametrically generated 2D Gridworld mazes.  For a given maze, our model predicts an optimal amount of partial planning at a state. For example, Figure~\ref{fig:exp1}a displays two mazes that our model predicts will have different immediate partial planning costs at the initial state. Since this cost reflects information processing, we can operationalize it in this task based on reaction times (RTs) as in previous work (e.g., \citealp{ortega2016}). Specifically, we assume that the amount of time between when a participant is presented with a maze and when they take their first action reflects the cost of encoding a partial plan from the initial state.

\subsection{Materials, Participants, and Procedure}
The stimuli were a set of 50 different $9 \times 9$ Gridworld mazes in which the start state was in the lower right corner and the goal state was in the upper left corner. These were chosen by first randomly generating a batch of 2,000 mazes and then selecting a random subset that were predicted by the model to have a range of different optimal planning costs at the initial state. 

We recruited 50 participants from Amazon Mechanical Turk and used the psiTurk framework~\cite{gureckis2016psiturk}. Each participant was paid a base pay of \$1.00. After reading the instructions and familiarizing themselves with the general mechanics of the task, participants started the main part of the task that included the 50 mazes. Each round, participants were first shown a blank $9 \times 9$ grid. When they pressed the spacebar, the maze for that round appeared immediately and they could move their agent (a blue circle) using the arrow keys. The initial-state RT measure was the time measured between the appearance of the maze on a round and their first action. When they reached the goal state, they received $+100$ points (50 points = 1\cent; total bonus = \$1.00). 

\begin{table*}[ht]
\centering
\begin{tabular}{rrrrr}
  \hline
 Predictor & LL Ratio [$\chi^2(1)$] & $p$ & $\beta$ & SE \\ 
  \hline
 { Partial Planning Cost } & 37.71 & $<.0001$ & 0.15 & 0.02 \\ 
  A* Node Count & 18.60 & $<.0001$ & 0.00 & 0.00 \\ 
  Optimal Plan Length & 14.06 & $<.001\ \ $ & -0.02 & 0.01 \\ 
  Information Theoretic Bounded Rationality & 0.82 & $0.36\ \ \ \ $ & 0.00 & 0.00 \\ 
  RL Softmax Entropy & 1.50 & $0.22\ \ \ \ $ & 0.32 & 0.26 \\ 
  Soft-Bellman Entropy & 10.38 & $<.001\ \ $ & -0.67 & 0.21 \\ 
  VI Iterations & 24.12 & $<.0001$ & 0.01 & 0.00 \\ 
  Trajectory Turns & 13.93 & $<.001\ \ $ & 0.02 & 0.00 \\ 
   \hline
\end{tabular}
\caption{Experiment 1 (Parametric Mazes) likelihood ratio tests and model estimates. Even when multiple planning metrics are included, the Partial Plan Cost derived from planning to plan is predictive of RTs.}
\label{table:exp1}
\end{table*}

\subsection{Planned Information Processing and Alternative Models}
In this experiment, we are interested in the minimized information theoretic planning cost at the initial state predicted by planned information processing. Formally, this corresponds to the $C(\tilde{\pi}^\beta, \bar\pi)$ term in Equation~\ref{eq:plantoplan} for an initial state $s_0$. These values were calculated for each maze using our algorithm and the same parameters as in the simulation.

To assess whether people are not simply planning, but adaptively planning their information processing, we considered seven alternative planning-based metrics as predictors. First, we considered the length of the shortest path from the start state to the goal, calculated using value iteration (Optimal Plan Length). Second, we ran A$^*$ search~\cite{hart1968formal}, a classical planning algorithm that finds a shortest path by maintaining a prioritized set of states to explore, starting with an initial state, and then iteratively exploring states and adding connected states to the exploration set until the goal state is reached. To facilitate better exploration, we provided A$^*$ with a Manhattan distance heuristic to the goal. We considered the number of candidate states explored by the algorithm before termination (A$^*$ Node Count). 

Third, we analyzed the action cost associated with the first step of the boundedly rational planning method proposed by~\citeauthor{Ortega2015}, \citeyear{Ortega2015} (Information theoretic bounded rationality). We set the information theoretic cost to be $\frac{1}{\alpha} = \lambda = .01$ to be commensurate with our planning to plan implementation. Fourth, we calculated the initial state entropy of a standard softmax over the optimal value function, with $\beta = 1$ (RL Softmax Entropy). Fifth, we calculated the initial state entropy of an optimal soft-Bellman policy, with $\beta = 1$ (Soft-Bellman Entropy). Sixth, we calculated the number of iterations of standard (planning) value iteration before convergence as a measure of planning computation (VI Iterations). Finally, as a heuristic measure of the complexity of a grid, we calculated the mean number of ``turns'' that occurred along a trajectory sampled from the optimal policy (Trajectory Turns).

If people are only planning actions, then the planning-based metrics should be sufficient for predicting RTs. If people are planning information processing---that is, constructing a computationally inexpensive partial plan that provides a good action at their current state---then the partial planning cost in our model would separately predict RTs.

\subsection{Results and Discussion}
We analyzed our data by comparing the predictions of the models to participants' initial-state RTs. Two participants had substantial missing data, and outliers were excluded, which left $2373$ initial-state RT measurements. To assess the relative predictive power of the eight models, we first fit a fully specified mixed effects linear model to log-normalized RTs. This included by-participant intercepts and round number slopes as random effects, and Partial Planning Costs as well as the seven planning-metrics as fixed effects. We then performed log-likelihood ratio tests with lesions versions that did not include each of the eight fixed effects. As shown in Table~\ref{table:exp1}, although several of the planning metrics are significant predictors, Partial Planning Cost not only predicts RTs, it is the predictor with the highest log-likelihood ratio test statistic. Thus, Experiment 1 suggests that people engage in planned information processing.

\section{Experiment 2: Probing Partial Plans}
Experiment 1 tested planning information processing using initial-state RTs to measure planning costs. In Experiment 2, we more directly examine whether people's partial planning is captured by our model. To probe partial plans, we used a technique of teleporting participants' avatars in Gridworld mazes and measuring their reactions. For instance, imagine that, in the middle of packing your bag for a flight you are unexpectedly teleported to the main terminal of the airport with your bag. It is likely you would quickly know what to do next (e.g., pick up your boarding pass) if that were part of your partial plan prior to being teleported. In contrast, if you had not thought that far ahead, then it would likely take you longer to determine your next action. Thus, \textit{post-teleportation reaction times} can be used to measure the divergence between a pre-teleportation plan and a post-teleportation plan.

\subsection{Materials, Participants, and Procedure}
The experiment consisted of 64 rounds of Gridworld mazes. To generate the mazes, four base $12 \times 12$ mazes were generated such that the initial state was in the lower right corner and the goal state was in the upper left corner. These were then transformed using the eight symmetries of a square, yielding a total of 32 perceptually distinct mazes. Each of the 32 mazes appeared twice. Half of the rounds were Normal rounds while the other half were Teleportation rounds. On the Teleportation rounds, a random number $n$ between 1 and the length of an optimal path for a maze was chosen, and on the $n$-th trial, the agent was hidden for 750ms and could not be controlled. It then reappeared in a randomly chosen location in the maze and could be controlled immediately. The amount of time between the reappearance of the circle and the participant's response was the post-teleportation RT measure that is the focus of this study. 

Sixty participants from MTurk were recruited for our experiment and given the same familiarization procedure as in Experiment 1. Participants were paid a base pay of \$1.00 and received a bonus of \$1.28 for completing the 64 trials.

\begin{table*}[!t]
\centering
\begin{tabular}{lrrrrrr}
\toprule
Predictor &  LL Ratio [$\chi^2(1)$] & $p$ & $\beta$ & SE \\
\midrule
Partial-Plan Divergence & $24.70$ & $<.001$ & $0.001$ &  $0.0003$ \\
A$^*$ Destination Nodes & $0.25$ & $.62\ \ $ & $-0.002$ & $0.005\ \ $ \\
A$^*$ Node Difference & $2.80$ & $.09\ \ $ & $-0.003$ &  $0.002\ \ $ \\
Optimal Path Length & $0.02$ & $.89\ \ $ & $0.001$ & $0.01\ \ \ \ $ \\
\bottomrule
\end{tabular}
\vspace{2mm}
\caption{Experiment 2 (Probing Partial Plans) likelihood ratio tests and model estimates.}
\label{table:exp3res}
\end{table*}

\subsection{Model-Predictor Variables}

Our goal is to explain post-teleportation RTs as a function of pre- and post-teleportation states and task structure. Our account provides partial plans at the pre- and post-teleportation states. If these map onto people's partial plans, then RTs will reflect a process of updating the pre-teleportation plan into the post-teleportation plan. To quantify this updating (i.e., re-planning) process, we calculated the state--action divergence between the pre-teleportation partial plan $\tilde\pi$ and post-teleportation partial plan $\tilde\pi'$, $D_{\text{KL}}[p^{\tilde\pi'}(a, s) || p^{\tilde\pi}(a, s)]$, where $p^\pi(a, s) = \pi(a \mid s) \rho^\pi(s)$. This ``Partial-Plan Divergence'' reflects the cost to encode the state--action distribution of the partial plan at the post-teleportation state starting from the one at the pre-teleportation state. It should thus reflect participants' ``new'' planning at the post-teleportation state. The same parameters as in the model in Experiment 1 were used to calculate the partial plans.

We calculated several alternative planning measures. First, we calculated the length of the optimal path from the post-teleportation state to the goal. Second, we calculated the number of A$^*$ nodes from the post-teleportation state (A$^*$ Destination Nodes). Third, we calculated an A$^*$ Node Difference score, corresponding to the additional nodes that A$^*$ explores at the post-teleportation state, taking into account those already explored at the pre-teleportation state.   

\subsection{Results and Discussion}
To assess the influence of the different predictors on post-teleportation log-normalized RTs, we used a similar mixed-effects linear model as in Experiment 1 (random effects were by-participant intercepts and round number, by-maze intercepts; fixed effects were model predictors). As summarized in Table~\ref{table:exp3res}, the Partial-Plan Divergence significantly predicted log-transformed RTs on post-teleportation trials. The planning-based predictors did not account for how quickly people reacted after being teleported. 

Additionally, we conducted a separate analysis that included teleportation distance as a fixed effect in the full model. Note that unlike the planning and partial planning models, teleportation distance is not an explicit model of decision-making. In this new model, partial planning is significant but weakened ($\chi^2(1) = 4.8$, $p < .05$). Additional details are included in the Supplementary Materials.

Thus, overall, the results of Experiment 2 suggest that people's RTs are explained by planned information processing via partial planning and not simply by planning actions.

\section{General Discussion}
\begin{figure}[b]
\centering
\includegraphics[width=.45\textwidth]{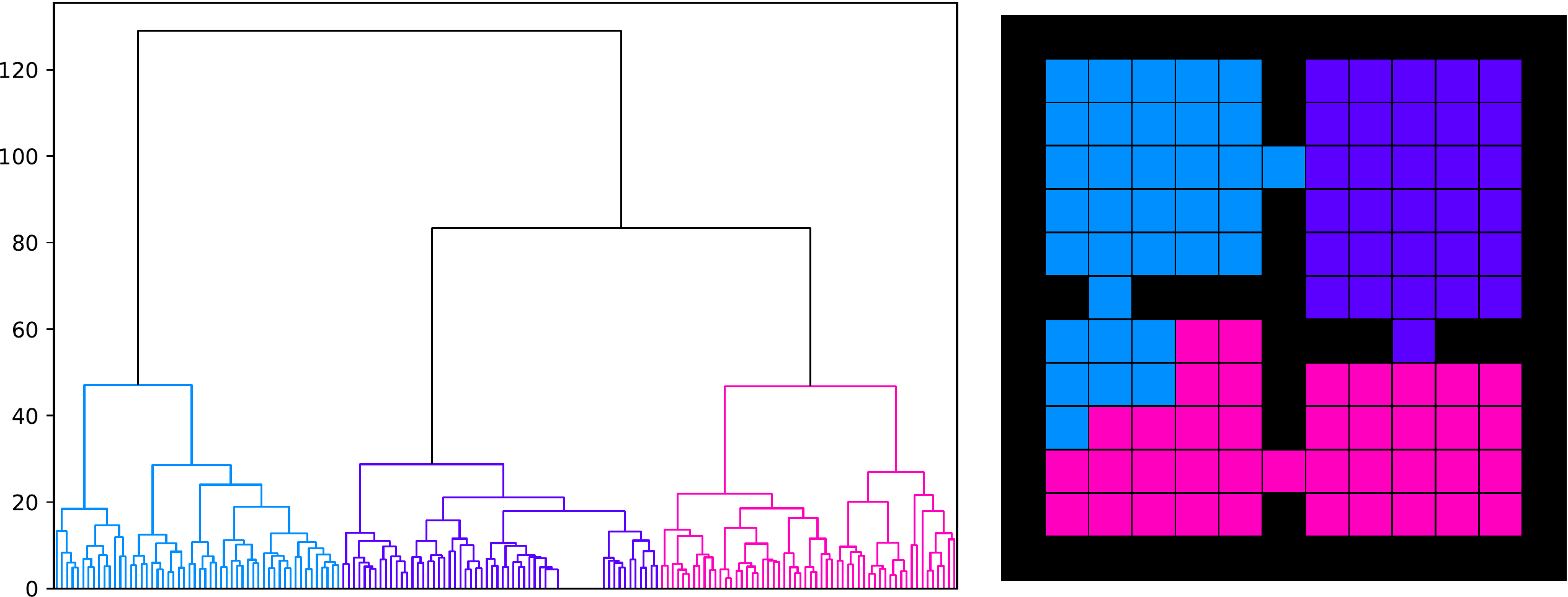}
\caption{States clustered by partial planning similarity resemble option representations.}
\label{fig:clustering}
\end{figure}

This paper asks two questions. First, why plan one's information processing? We argue that meta-planning lets agents adaptively capitalize on the benefits of planning while regulating planning costs. To make this precise, we formalize the general notion of \textit{partial plans} that prioritize planning in different parts of a simulated model and define an \textit{information-theoretic encoding cost} for partial plans, enabling us to define a novel recursive Bellman objective that includes both task rewards and planning costs. This model provides a point of departure for future normative accounts of human meta-planning.

Second, do people plan their information processing? We reported two human experiments that test our formal account of planned information processing. Experiment~1 demonstrates that adaptive partial planning explains people's initial reaction times when navigating parametrically generated 2D mazes. Experiment~2 used unexpected teleportations while navigating mazes to probe partial planning representations. The optimal partial plans generated by our model explain human responses even when accounting for action planning.

People plan because planning is useful. But, planning is hard, so people make planning easier by being selective about what and when they plan. In other words, people should plan their planning. For the most part, current decision-making algorithms plan, albeit with the help of good heuristics and abstractions provided by computer scientists. But, ideally, algorithms would learn how to make planning easier for themselves by planning their planning. 
Understanding planned use of computational resources can also provide insight into the nature and function of abstractions when \textit{learning}. For instance, we performed hierarchical clustering over states in Four Rooms based on the similarity of their optimal partial plans (Figure~\ref{fig:clustering}; details in supplementary materials), which results in clusters resembling options from research on hierarchical reinforcement learning~\cite{sutton1999between,dietterich2000hierarchical,parr1998reinforcement,Botvinick2008}.
In short, this work is an important step towards understanding the scale and sophistication of human meta-planning and applying such insights to the design of machines.

\section*{Acknowledgements}
The authors would like to thank Daniel Reichman, Bill Thompson, Fred Callaway, and Rachit Dubey for their advice and feedback on this work. This research was supported by NSF grant \#1544924, AFOSR grant \#FA9550-18-1-0077, and grant \#61454 from the John Templeton Foundation.

\bibliographystyle{aaai}
\bibliography{bibliography}



\onecolumn
\section*{Supplemental Materials}
\subsection*{Solving Planning to Plan}
\newcommand{\allsaz}{\;\;\;\;\;\;\forall a, s, \tilde{s}}
\newcommand{\allsz}{\;\;\;\;\;\;\forall s, \tilde{s}}
\newcommand{\alls}{\;\;\;\;\;\;\forall s}

\begin{algorithm}
    \LinesNumbered
  \SetAlgoLined
  \caption{Gradient-based Planning to Plan}
  \BlankLine
\SetKwInOut{Input}{Input}
    \SetKwInOut{Output}{Output}
    \Input{MDP $M$, Planning cost weight $\lambda$, Default policy $\bar\pi$, \newline
    Planning to Plan Iterations $N$, Partial Plan Horizon $H$}
    \Output{Temperature allocation $\beta$, Partial Plans $\tilde\pi^{\beta^*}$}
  \BlankLine
  $\tilde{S} \leftarrow S$\;
  \tcc{Initialize temperatures}
  $\beta_0(\tilde{s} ; s) \leftarrow$ \text{Randomly initialize for all } $s \in S, \tilde{s} \in \tilde{S}$\;
  \BlankLine
  \tcc{Planning to plan loop}
  \For{$n=0$ \KwTo $N$}{
  \BlankLine
  \tcc{Partial planning loop}
   $\tilde{q}^{\beta_{n}}_0(\tilde{s}, a; s) \leftarrow 0\allsaz$\;
   \For{$t = 0$ \KwTo $H$}{
   	$\tilde{\pi}^{\beta_n}_t(a \mid \tilde{s} ; s) 
       \propto \exp \big\{\tilde{q}^{\beta_n}_t(\tilde{s}, a; s)\beta_n(\tilde{s}; s)\big\}\allsaz$\;
       $\tilde{v}^{\beta_{n}}_{t}(\tilde{s};s) \leftarrow 
       \sum_a \tilde{\pi}^{\beta_n}_t(a \mid \tilde{s} ; s)\tilde{q}^{\beta_n}_t(\tilde{s}, a; s)\allsz$\;
       $\tilde{q}^{\beta_n}_{t+1}(\tilde{s}, a; s) \leftarrow 
       \sum_{\tilde{s}'}T_{\tilde{s},\tilde{s}'}^a [
       R_{\tilde{s},\tilde{s}'}^a + \gamma \tilde{v}^{\beta_{n}}_{t}(\tilde{s};s)
       ]\allsaz$
   }
   \BlankLine
   \tcc{Calculate information theoretic costs of partial plans}
   $c_n(\tilde{s}; s) \leftarrow 
     D_{\text{KL}}\big[\tilde{\pi}^{\beta_n}_H(\cdot \mid \tilde{s} ; s) || \bar\pi(\cdot \mid \tilde{s})\big]
     \allsz$\;
   $c_n(s) \leftarrow \sum_{\tilde{s}}c_n(\tilde{s}; s)\alls$\;
   \BlankLine
   \tcc{Calculate Planning to Plan Value (i.e. Policy Evaluation)}
   $V_\lambda^{\beta_n}(s) \leftarrow \sum_a 
   \Big[ \tilde{\pi}^{\beta_n}_H(a \mid s; s)\sum_{s'}
   T_{s, s'}^a \big[R_{s, s'}^a + \gamma V_\lambda^{\beta_n}(s')\big]
   \Big]
   - \lambda c_n(s)
   \alls
   $\;
   \BlankLine
   \tcc{Calculate Loss and Gradient, then Update Temperatures}
   $L \leftarrow -\sum_{s}{V_\lambda^{\beta_n}(s)}$\;
   $\delta_n \leftarrow \frac{\partial L}{\partial \beta_n}$\;
   $\beta_{n+1} \leftarrow \text{Adam}(\beta_n, \delta_n)$
}
$\beta^* \leftarrow \beta_N$\;
$\tilde{\pi}^{\beta^*} \leftarrow \tilde{\pi}^{\beta_N}_H$\;
\KwRet{$\beta^*, \tilde{\pi}^{\beta^*}$}
\label{alg:gradientbased}
\end{algorithm}

Solving the planning-to-plan objective allows us to understand its qualitative features as well as derive predictions of human decision-making. Algorithm~1 describes a gradient-based procedure for solving this problem, and conveys the main ideas of our account procedurally. The output of an inner loop of partial planning (lines 4 to 9) is evaluated based on the information-theoretic cost of plans (lines 10 and 11) and sequential decision-making rewards (line 12). This is nested within an outer Planning to Plan loop (lines 3 to 16) that optimizes the partial plans.


\subsection*{Deriving Option-like Representations}

Our account concerns planning, but also sheds light on the nature of representations that facilitate good learning. We examined how option-like representations can emerge from the optimal partial planning process by clustering states based on their planning similarity.

\begin{figure}[!hb]
\centering
\includegraphics[width=\textwidth]{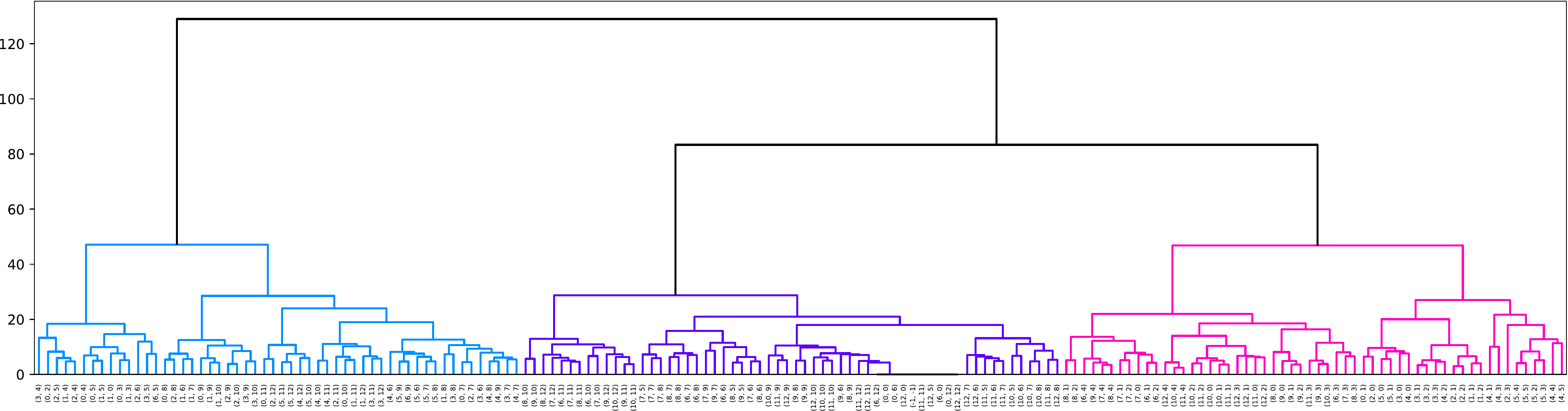}
\caption{State dendrogram based on planning similarity in Four Rooms example}
\label{fig:dendrogram}
\end{figure}

\begin{figure}[!hb]
\centering
\includegraphics[width=.3\textwidth]{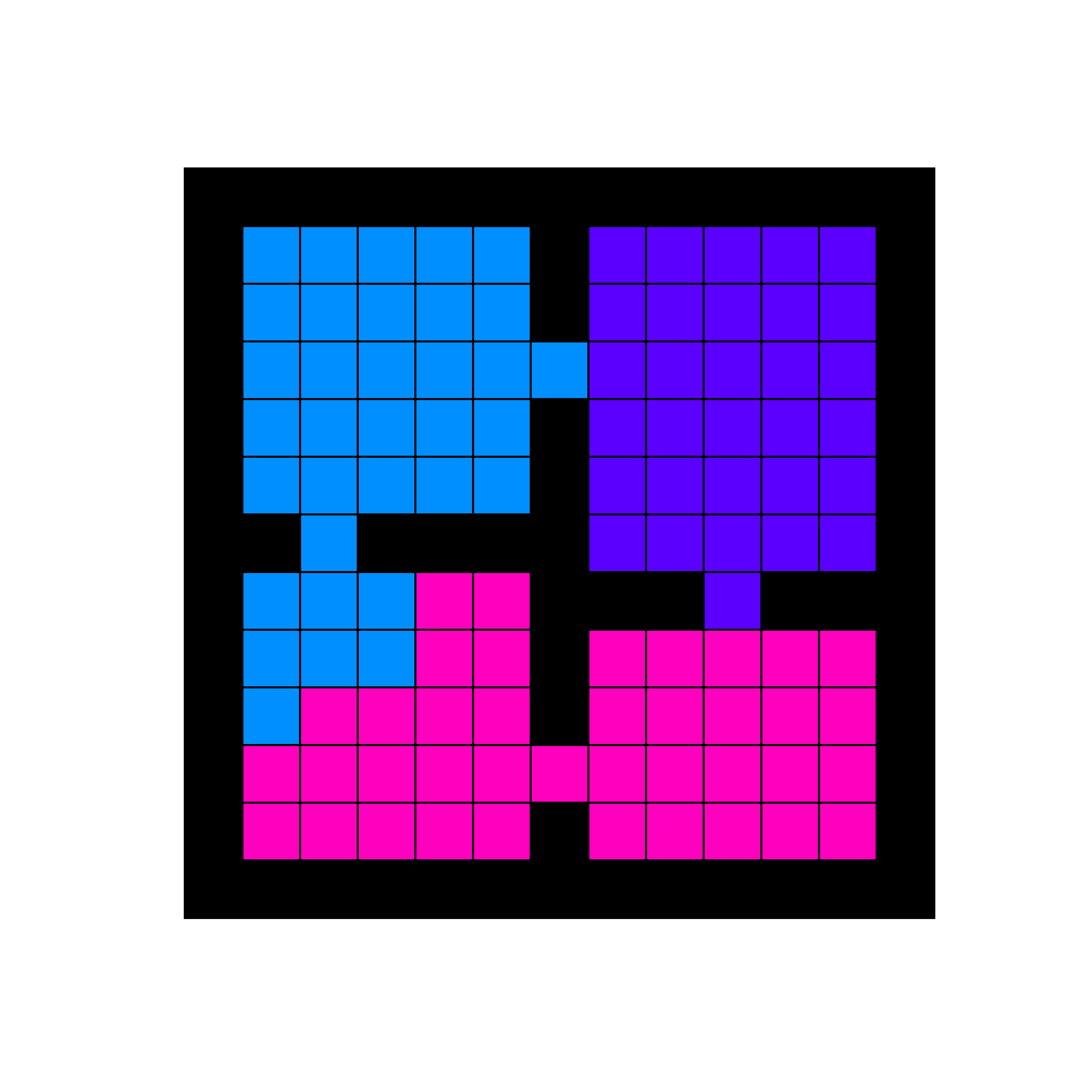}
\caption{Top three clusters.}
\label{fig:options}
\end{figure}

To examine how the plans at different states in Four Rooms relate to one another, we calculated a \textit{symmetric planning distance} for each pair of states. Specifically, for each pair of ground states $s_A$ and $s_B$, we calculated 
\begin{equation}
    d(s_A, s_B) = 
\sum_{\tilde{s}\in \tilde{S}}
D_{\text{KL}}[\tilde{\pi}^{\beta^*}(\cdot \mid \tilde{s} ; s_A) || \tilde{\pi}^{\beta^*}(\cdot \mid \tilde{s} ; s_B) ] + 
D_{\text{KL}}[\tilde{\pi}^{\beta^*}(\cdot \mid \tilde{s} ; s_B) || \tilde{\pi}^{\beta^*}(\cdot \mid \tilde{s} ; s_A) ].
\end{equation}
We then performed hierarchical clustering using Ward's method with the distance matrix $d$. Figure~\ref{fig:dendrogram} shows the results of hierarchical clustering. Figure~\ref{fig:options} shows the largest three clusters. The first cluster is the room containing the goal, while the other two clusters each contain one of the two intermediate rooms and half of the starting room. Although our goal here is to better understand \textit{meta-planning}, we note that these clusters are highly reminiscent of options used in hierarchical reinforcement learning.

\subsection*{Experiment 2 Additional Analysis}
\begin{table*}[!ht]
\centering
\begin{tabular}{lrrrrrr}
\toprule
Predictor &  LL Ratio [$\chi^2(1)$] & $p$ & $\beta$ & SE \\
\midrule
Partial-Plan Divergence & $4.79$ & $<.05$ & $0.0007$ &  $0.0003$ \\
A$^*$ Distance to Goal & $0.18$ & $.67$ & $-0.002$ & $0.005$ \\
Teleportation Distance & $15.83$ & $<.001$ & $0.01$ & $0.003$ \\
A$^*$ Node Difference & $5.38$ & $<.05$ & $-0.004$ &  $0.002$ \\
Optimal Path Length & $0.23$ & $.63$ & $0.005$ & $0.01$ \\
\bottomrule
\end{tabular}
\vspace{2mm}
\caption{Experiment 2 (Probing Partial Plans) likelihood ratio tests and model estimates where full model includes teleportation distance.}
\label{table:exp3res_secondary}
\end{table*}

For Experiment 2, we conducted a secondary analysis in which the full mixed-effects linear model included the partial-plan divergence, A$^*$ Distance to Goal, A$^*$ Node difference, optimal path length, and teleportation distance. Unlike the other metrics, which are based on a planning model, the teleportation distance metric is derived by taking the Euclidean distance between the pre-teleportation and post-teleportation state. In this model, teleportation distance is strongly predictive, suggesting that future work is needed to disentangle the contributions of planning, meta-planning, and other processes in predicting human RTs.

\subsection*{Experimental Stimuli}
\begin{figure}[h!]
\centering
\includegraphics[width=.7\textwidth]{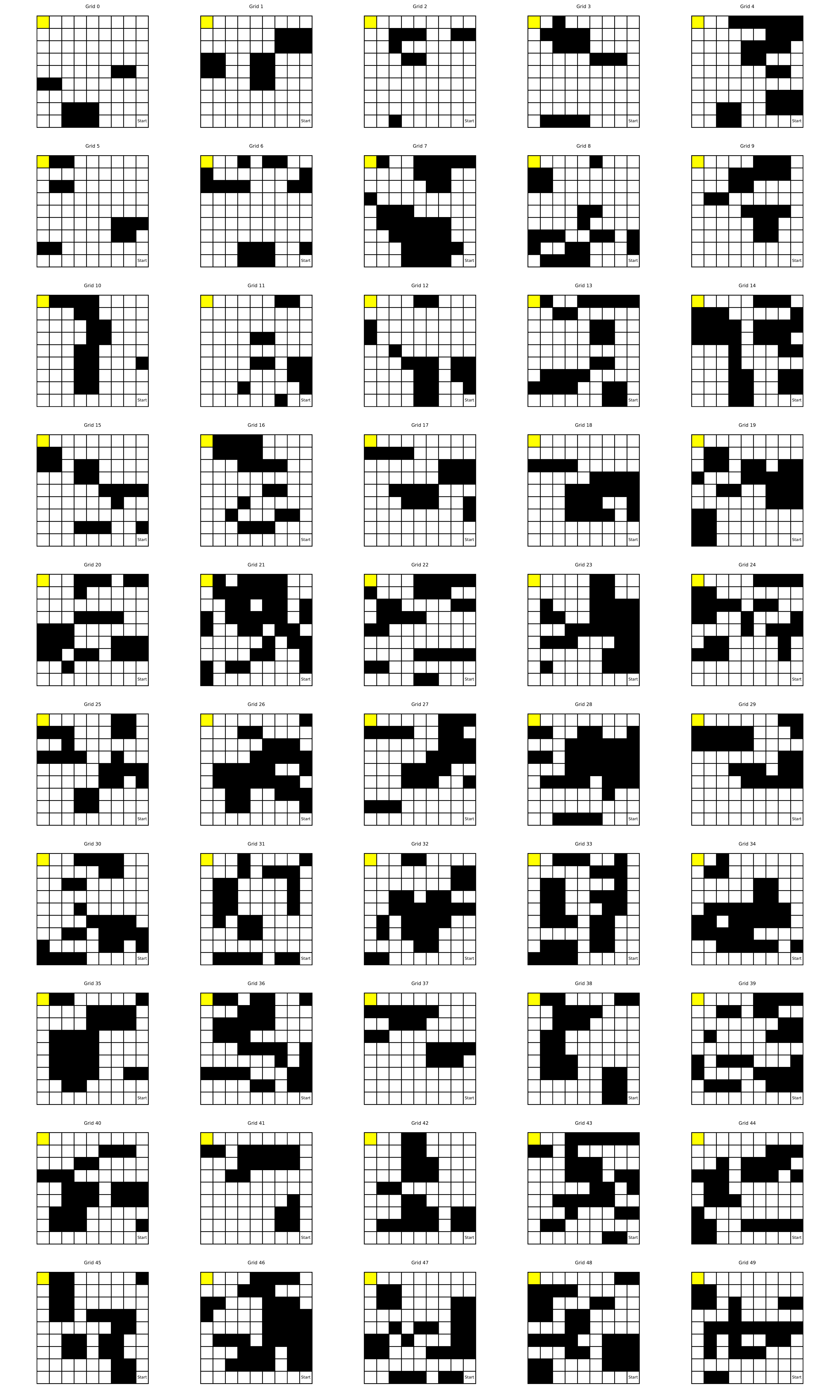}
\caption{Mazes used in experiment 1.}
\label{fig:}
\end{figure}

\begin{figure}[h!]
\centering
\includegraphics[width=\textwidth]{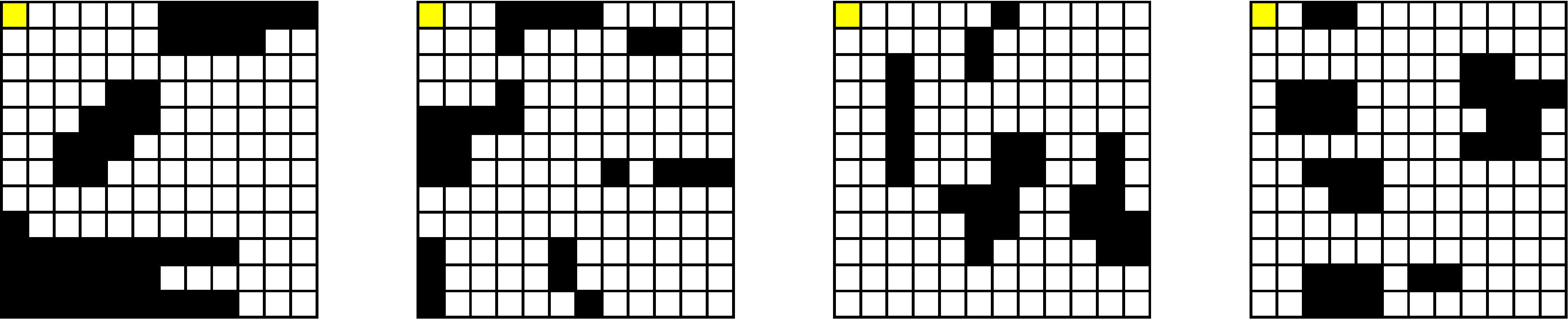}
\caption{Mazes used in experiment 2. Start state was in the lower right. Note that the complete set of stimuli included eight transformations of these mazes, resulting in 32 unique mazes, each of which appeared once as a Teleportation round and once as a Normal round.}
\label{fig:}
\end{figure}


\end{document}